\documentclass{article}

\usepackage{PRIMEarxiv}
\usepackage{tabularx} 
\usepackage{amsmath}
\usepackage[utf8]{inputenc} 
\usepackage[T1]{fontenc}    
\usepackage{hyperref}       
\usepackage{url}            
\usepackage{booktabs}       
\usepackage[utf8]{inputenc}

\usepackage{amsfonts}       
\usepackage{nicefrac}       
\usepackage{microtype}      
\usepackage{lipsum}
\usepackage{fancyhdr}       
\usepackage{graphicx}       
\usepackage{hyperref}

\usepackage[utf8]{inputenc}
\usepackage[arabic,farsi,english]{babel}

\usepackage{listings}
\usepackage{xcolor}
\usepackage{tcolorbox}

\graphicspath{{media/}}     

\title{Persian-Phi: Efficient Cross-Lingual Adaptation of Compact LLMs via Curriculum Learning
 
}

\author{
  Amir Mohammad Akhlaghi \\
  Shahid Beheshti University \\
  Tehran, Iran \\
  \texttt{a.akhlaghigharelar@mail.sbu.ac.ir} \\
   \And
  Amirhossein Shabani \\
  Shahid Beheshti University \\
  Tehran, Iran \\
  \texttt{am.shabani@mail.sbu.ac.ir} \\
   \And
  Mostafa Abdolmaleki \\
  Shahid Beheshti University \\
  Tehran, Iran \\
  \texttt{mos.abdolmaleki@mail.sbu.ac.ir} \\
   \And
  Saeed Reza Kheradpisheh$^*$ \\
  Shahid Beheshti University \\
  Tehran, Iran \\
  \texttt{s\_kheradpisheh@sbu.ac.ir} \\
}

\begin{document}
\maketitle

\renewcommand{\thefootnote}{\fnsymbol{footnote}}
\footnotetext[1]{Corresponding Author}

\begin{abstract}
Abstract The democratization of AI is currently hindered by the immense computational costs required to train Large Language Models (LLMs) for low-resource languages. This paper presents Persian-Phi, a 3.8B parameter model that challenges the assumption that robust multilingual capabilities require massive model sizes or multilingual baselines. We demonstrate how Microsoft’s Phi-3 Mini—originally a monolingual English model—can be effectively adapted to Persian through a novel, resource-efficient curriculum learning pipeline. Our approach employs a unique "warm-up" stage using bilingual narratives (Tiny Stories) to align embeddings prior to heavy training, followed by continual pretraining and instruction tuning via Parameter-Efficient Fine-Tuning (PEFT). Despite its compact size, Persian-Phi achieves competitive results on \href{https://huggingface.co/spaces/PartAI/open-persian-llm-leaderboard}{Open Persian LLM Leaderboard}. Our findings provide a validated, scalable framework for extending the reach of state-of-the-art LLMs to underrepresented languages with minimal hardware resources. The model is publicly available at \href{https://huggingface.co/amirakhlaghiqqq/PersianPhi}{Persian-Phi}.

\end{abstract}


\section{Introduction}

LLMs exhibit significant performance disparities across languages. When queried in languages other than those they were primarily trained on—such as English or Chinese—their reasoning, comprehension, and problem-solving capabilities degrade noticeably \cite{aryabumi2024aya23openweight, jiang2024mixtralexperts}. This linguistic gap means that non-native speakers of high-resource languages often experience AI at a significantly lower capacity, limiting their access to cutting-edge advancements.

A straightforward solution is to train a separate model from scratch for each language. However, this approach is hindered by two major challenges:

\begin{enumerate}
    \item Data scarcity: Classic data scaling laws suggest that fewer than 15 languages (excluding Persian) have sufficient high-quality and diverse datasets to train robust LLMs \cite{common_crawl, weber2024redpajamaopendatasettraining}.
    \item  Computational cost: Training large-scale models requires immense hardware resources, which are often infeasible for communities of many underrepresented languages.
\end{enumerate}

To address this, some researchers have explored continued pretraining of multilingual models on a target language to improve performance\cite{cui2023efficient}. However, this method comes with structural limitations. Multilingual models distribute their fixed parameter capacity across multiple languages, leading to capacity competition and diluted performance. Empirical findings show that monolingual models—when trained under similar conditions and model sizes—often outperform their multilingual counterparts in the target language\cite{chang2023multilingualitycurselanguagemodeling}.

This paper proposes an alternative approach: rather than relying on a multilingual base model, we start with a high-capability monolingual English model and strategically enrich it with Persian language capabilities. This method preserves the advantages of monolingual specialization while enhancing cross-lingual adaptability, offering a scalable and resource-efficient solution for low-resource languages. Our key contributions include

\begin{itemize}
\item A novel cross-lingual curriculum learning pipeline: (1)  Initial embedding alignment of an extended tokenizer via translation tasks, (2) Continual pretraining on rigorously filtered Persian corpora (TLPC, Wikipedia) and Tiny Stories for knowledge transfer, and (3) Bilingual instruction fine-tuning using LoRA to balance Persian fluency with English retention.

\item A Persian LLM with 3.85B parameters that achieves competitive performance against other similarly sized multilingual LLMs supporting Persian on key benchmarks.

\item Creation of a high-quality Persian language dataset through a rigorous data processing pipeline, including content filtering, normalization, de-duplication, and quality control across diverse sources.

\end{itemize}

Our results demonstrated that fine-tuning Microsoft’s Phi-3 Mini—originally an English-only model—can effectively extend its capabilities to Persian. Unlike multilingual models that distribute capacity across many languages, our approach focuses adaptation on a single target language, making more efficient use of the model’s parameters. This strategy provides a practical and scalable pathway for enabling support for other low-resource languages after pretraining, without modifying the model’s architecture or requiring extensive retraining.


\section{Related Works}
\subsection{Efficient LLMs and the Phi Series}

For years, the development of LLMs followed the principle that "bigger is better." Early models like GPT (117M)\cite{Radford2018ImprovingLU}, GPT-2 (1.5B)\cite{Radford2019LanguageMA}, and GPT-3 (175B)\cite{brown2020languagemodelsfewshotlearners} showed consistent gains through scale, particularly in zero- and few-shot learning. Open-source models such as OPT (125M–175B)\cite{zhang2022optopenpretrainedtransformer} and LLaMA (1B–405B)\cite{touvron2023llamaopenefficientfoundation, touvron2023llama2openfoundation, grattafiori2024llama3herdmodels} reinforced this trend.

However, the emergence of Microsoft’s Phi series\cite{gunasekar2023textbooksneed} challenged this mindset. Phi-1 (1.3B), trained on a carefully curated mix of filtered web data and synthetic GPT-3.5 textbooks, outperformed much larger models, achieving strong results on HumanEval (50.6\%) and MBPP (55.5\%)\cite{chen2021evaluating, austin2021program}. Subsequent iterations—Phi-2 and Phi-3\cite{abdin2024phi3technicalreporthighly}—further refined data quality and reduced dependence on synthetic sources. Phi-3 Mini (3.8B) reached performance comparable to GPT-3.5, aided by architectural improvements like LongRope\cite{ding2024longropeextendingllmcontext} for extended context.
The family recently includes Phi‑4 with expanded multilingual support (200K tokens)\cite{abdin2024phi}.

By prioritizing data efficiency over size, the Phi models challenged the long-standing belief that larger models are always better, showing that carefully curated training data can unlock exceptional performance without requiring massive parameter counts. Based on the available computational resources and the demonstrated capabilities of Phi-3 in achieving impressive results with relatively modest parameter counts, we selected it as the base model for fine-tuning to Persian.

\subsection{Adapting LLMs to a New Language}
Some methods align LLMs to a new language by leveraging high-resource languages as a guide. For instance, Cross-lingual instruction-tuning\cite{zhu2023extrapolatinglargelanguagemodels} combines translation data and instruction-following tasks across languages to enhance cross-lingual capabilities. Pivot Language Guided Generation (PLUG)\cite{zhang2024plugleveragingpivotlanguage} first generates responses in a pivot language before translating them into the target language. Similarly, multilingual feedback techniques\cite{feng2024teachingllmsabstainlanguages, kuulmets2024teachingllamanewlanguage} train models to evaluate responses in multiple languages, learning when to abstain from answering. While these alignment-based strategies help direct LLMs toward lower-resource languages, they often rely heavily on high-resource languages (e.g., English). This can introduce biases, reduce fluency, and limit expressiveness in languages that differ significantly from the pivot language\cite{chen2024instructioncpfastapproachtransfer}.

Other approaches focus on directly adapting LLMs to new languages, aiming to improve both instruction-following and generative abilities. Instruction Continual Pre-training (InsCP)\cite{chen2024instructioncpfastapproachtransfer} refines models on target-language data while preserving safety and conversational consistency through special instruction tags. Efficient adaptation methods\cite{csaki2023efficientlyadaptingpretrainedlanguage} use tokenizer replacement to introduce new tokens without degrading performance in the original language, while data mixing prevents catastrophic forgetting. Mini-model adaptation\cite{marchisio2023minimodeladaptationefficientlyextending} trains new embeddings on a smaller model before integrating them into the full model, improving efficiency. Additionally, vocabulary expansion and supervised fine-tuning (SFT)\cite{chen2024instructioncpfastapproachtransfer, csaki2023efficientlyadaptingpretrainedlanguage} further enhance the model’s ability to generate fluent text while maintaining instruction-following skills across multiple languages.

\subsection{Persian LLMs}

The development of Persian LLMs has gained momentum, addressing the limitations of early open-source models trained primarily on English data. PersianMind \cite{rostami2024persianmindcrosslingualpersianenglishlarge}, an open-source bilingual Persian-English LLM, exemplifies this progress. Built on the LLaMa 2 architecture, it expands the vocabulary with 10,000 Persian subwords and undergoes training on a 2 billion Persian token corpus. Employing Low-Rank Adaptation (LoRA) for efficient fine-tuning, PersianMind achieves comparable performance to GPT-3.5-turbo in Persian reading comprehension and tries to mitigate English knowledge forgetting through parallel dataset training. Similarly, the Maral model\cite{maralgpt2023maral7b}, a 7 billion parameter bilingual model based on the Mistral architecture, leverages a translated Alpaca dataset for fine-tuning.


More recently, the Dorna family of models\cite{partai2024dorna_llama3_8b}, particularly Dorna-Llama3-8B-Instruct built upon Meta Llama 3 Instruct, showcases further advancements. Evaluations indicate Dorna's superior performance on Persian tasks compared to its base model and its competitiveness against larger models and closed-source alternatives. These models utilize techniques like LoRA and fine-tuning on closed-source datasets to enhance their Persian language capabilities. However, challenges remain, including mitigating hallucination and catastrophic forgetting, and developing larger, higher-quality Persian datasets.

\begin{figure}
\centering
\includegraphics[width=1\linewidth]{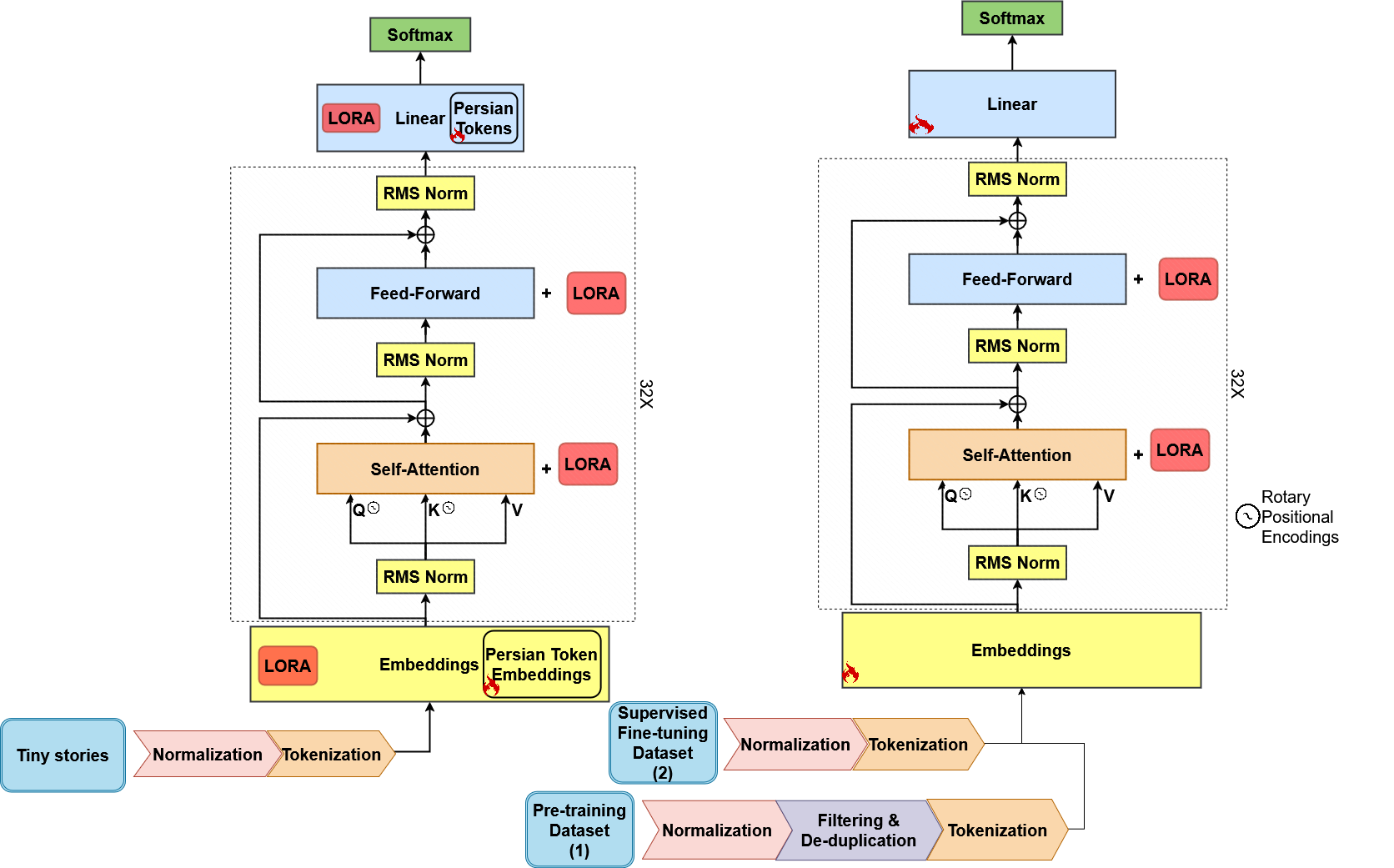}
\caption{The cross-lingual adaptation pipeline began with a warm-up phase (left), which introduced an extended tokenizer with new Persian tokens. Using translated Tiny Stories, this stage employed low-rank LoRA fine-tuning alongside full fine-tuning of the new embedding and head parameters to smoothly initialize and align these tokens. The main adaptation phase (right) involved continual pre-training on a large, filtered and deduplicated Persian corpus, utilizing higher-rank LoRA on attention/feed-forward layers plus full embedding/head tuning to build deep language understanding. Finally, supervised fine-tuning (SFT) was performed on a mixed instruction dataset (Persian, English), again using LoRA, to refine instruction-following and conversational abilities in both languages, ensuring Persian proficiency was added while retaining original English capabilities.}
\label{fig:enter-label}
\end{figure}

\section{Methodology}
In this section, we detail our methodology for adapting the Phi-3-mini-instruct language model to effectively process and generate Persian text while preserving its original capabilities, particularly in English. Our approach consisted of four key stages: (1) Tokenizer Preparation, where we enhanced the existing LLaMA-2 tokenizer by incorporating Persian-specific tokens to address the language's morphological complexity; (2) New Parameters Warm-up, which employed a bidirectional translation task to integrate new tokens smoothly and mitigate challenges such as catastrophic forgetting; (3) Pre-training, which involved the curation and processing of a large Persian dataset to ensure robust language understanding; and (4) Supervised Fine-Tuning, which used instruction-response pairs to reinforce the model's conversational abilities in both Persian and English. Each stage built upon the previous one, ensuring a comprehensive and effective adaptation process tailored to the unique linguistic demands of Persian.
\subsection{Tokenizer Preparation}

Effective tokenization was a crucial pre-processing step that significantly impacted language model performance. The base model in our study, Phi-3-mini-instruct, employed the LLaMA-2 tokenizer, which presented limitations when processing Persian text. While it supported basic Persian characters, it lacked specialized tokens for Persian-specific linguistic constructs and failed to efficiently handle Persian's rich morphological structure, which included extensive use of prefixes, suffixes, and compound words.

To address these limitations, we implemented a systematic approach to adapt the LLaMA-2 tokenizer for Persian language processing. Using the Persian Wikipedia corpus as our primary dataset, we normalized the text and trained a new Byte-Pair Encoding (BPE) model \cite{sennrich2015neural} with a vocabulary size of 5,000 tokens on the normalized corpus.\footnote{Due to character set overlap between the LLaMA-2 tokenizer and our newly trained tokenizer, 4,921 unique new tokens were actually added to the base model.} This process focused on identifying frequent subword patterns specific to Persian morphology, resulting in the addition of new Persian-specific tokens.

The vocabulary size was empirically determined through manual inspection to optimize the balance between morphological pattern coverage and computational efficiency, primarily by reducing sequence lengths for Persian texts. These improvements laid the foundation for more effective Persian language modeling in subsequent fine-tuning stages, while preserving the original capabilities of the model for other languages.

\begin{table*}
    \centering
    \caption{Comparison of enhanced tokenizer performance for a sample sentence. In this example, the enhanced tokenizer more effectively captured Persian language constructs such as prefixes and suffixes, whereas the base tokenizer processed the text at the character level. The enhanced tokenizer also reduced the overall number of tokens}
    \begin{tabular}{|p{0.45\textwidth}|p{0.45\textwidth}|}
        \hline
        \multicolumn{1}{|c|}{\textbf{LLaMA2 Tokenizer}} & \multicolumn{1}{c|}{\textbf{Enhanced Tokenizer}} \\
        \hline
        \FR{[ع , ل , ی ,  , ب , ا ,  , د , و , س , ت , ا , ن , ش ,  , ب , ه ,  , م , د , ر , س , ه ,  , ر , ف , ت , ن , د , .]} & 
        \FR{[علی ,  , با ,  , دوستان , ش ,  , به ,  , مدرسه,  , رفت , ند, .]} \\
        \hline
        \multicolumn{2}{|c|}{Original Sentence: \FR{ علی با دوستانش به مدرسه رفتند. }} \\
        \hline
        Number of tokens: 30 & Number of tokens: 14 \\
        \hline
    \end{tabular}
    \label{tab:tokenizer-comparison}
\end{table*}

\subsection{New Parameters Warm-up}

Introducing new tokens into a pre-trained language model posed significant challenges. Specifically, the embedding and language modeling head layers of the Phi-3-mini-instruct model had to be resized to accommodate the extended tokenizer. The new parameters introduced during this process were randomly initialized, which often made it difficult for the model to align the new tokens with its existing knowledge. Consequently, the model struggles to effectively leverage its prior understanding when learning a new language.

Furthermore, random initialization increases the risk of catastrophic forgetting, as the model requires extended fine-tuning, potentially disrupting its original capabilities. This issue has also been highlighted in recent studies—such as \cite{zhao2024llama}—which question the efficacy of extending tokenizers for improving cross-lingual capabilities without appropriate initialization strategies.

The primary challenges in extending pre-trained models to new languages through tokenizer modification include:

\begin{itemize}
    \item \textbf{Semantic Discontinuity:} Random initialization of new token embeddings creates a semantic gap between the pre-existing and newly added tokens, hindering the model's ability to leverage its pre-trained knowledge effectively.
    \item \textbf{Catastrophic Forgetting:} The model requires substantially more training steps to establish meaningful connections between randomly initialized parameters and existing knowledge, increasing the risk of catastrophic forgetting.
\end{itemize}

\subsubsection{Methodology}
To address the challenge of integrating newly introduced Persian tokens, we proposed a warm-up phase using a bidirectional translation task based on the Tiny-Stories dataset. By fine-tuning the model on aligned English–Persian sentence pairs, it gradually learned to incorporate Persian while preserving its existing English capabilities, mitigating the risk of catastrophic forgetting.

This structured exposure to simple narratives helps the model internalize key syntactic and semantic differences between the two languages. As a result, the embedding space is updated coherently, ensuring smoother cross-lingual alignment and laying a solid foundation for subsequent Persian-specific fine-tuning.
\subsubsection{Warm-up Dataset Preparation}

Effective alignment between English and Persian requires a high-quality, large-scale parallel corpus, which is lacking in publicly available datasets. To address this limitation, we constructed our own dataset by leveraging machine translation to generate a substantial parallel corpus. The Tiny-Stories dataset \cite{eldan2023tinystories}, a synthetic collection of short stories generated by GPT-4 with vocabulary and grammar suitable for a child aged 3–4 years, was selected for this stage due to its simplistic structure. This simplicity ensures better token alignment and facilitates the gradual adaptation of the model to Persian, replicating the process of learning a new language from basic concepts before progressing to more complex structures.

The dataset preparation process involved translating each story from English to Persian using the Google Translate API. Due to the simplicity of the stories, the machine translation achieved a high success rate with only minor errors. The translated stories were then organized in a parallel format, with alternating English and Persian sentences. This arrangement enabled the model to effectively compare and learn patterns between the two languages. Templates of this format are shown below:

\begin{center} 
\tcbset{colback=gray!10, colframe=gray!50, width=0.8\textwidth, boxrule=0.5mm, sharp corners}
\begin{tcolorbox}[title= Templates of training samples for the warm-up stage]
\begin{minipage}[t]{0.48\textwidth} 

<FA> [First paragraph in Persian] \\
<EN> [First paragraph in English] \\[1ex]
\hfill \textbf{...} \hfill \\[1ex] \\[1ex]
<FA> [Last paragraph in Persian] \\
<EN> [Last paragraph in English]
\end{minipage}%
\hfill
\vrule width 0.5mm 
\hfill
\begin{minipage}[t]{0.48\textwidth} 

<EN> [First paragraph in English] \\
<FA> [First paragraph in Persian] \\[1ex]
\hfill \textbf{...} \hfill \\[1ex] \\[1ex]
<EN> [Last paragraph in English] \\
<FA> [Last paragraph in Persian]

\end{minipage}

\end{tcolorbox}
\end{center}

\subsubsection{Warm-up Training Process}

The training process began by resizing the embedding and head layers of the Phi-3-mini-instruct model to accommodate the newly added Persian tokens, which were initialized randomly. The model was fine-tuned using LoRA with a rank of 4, a deliberate choice aimed at introducing minimal changes to the model’s weights to avoid catastrophic forgetting during fine-tuning. This configuration introduced approximately 6 million new parameters, striking a balance between computational efficiency and the capacity to adapt effectively. Furthermore, the head and embedding parameters of the newly added Persian tokens underwent full fine-tuning, ensuring that their embeddings were thoroughly integrated with the pre-existing linguistic framework of the model. The training centered on a next-token prediction objective, using approximately 500 million tokens from the Tiny-Stories dataset. Hyperparameters such as batch size and learning rate were kept consistent with the model’s original pre-training settings in ~\ref{sec:training_configurations}.

\subsubsection{Warm-up Outcome}
The effectiveness of our warm-up methodology was validated through both quantitative and qualitative analyses. Quantitatively, the model achieved a low perplexity score of 2.45 on the bilingual validation set. A perplexity of 2.45 corresponds to an average predicted probability of approximately 40.8\% for the correct next token, demonstrating that the model is confidently and accurately modeling the token sequences in the warm-up corpus—a strong indication that the new Persian embeddings were effectively aligned. This successful initial alignment was crucial, providing strong evidence that the warm-up phase achieved its primary objective: preparing the model for deep, continual pre-training without the instability often caused by randomly initialized parameters.

Qualitative analysis further supported these findings. Generated texts were assessed for grammatical accuracy, coherence, and narrative quality. The model demonstrated an ability to produce bilingual stories with appropriate syntactic and semantic structures. An example of a generated story is provided in Appendix \ref{tab:story_generation_example}.

\begin{figure}[h!]
\centering
\includegraphics[width=0.5\textwidth]{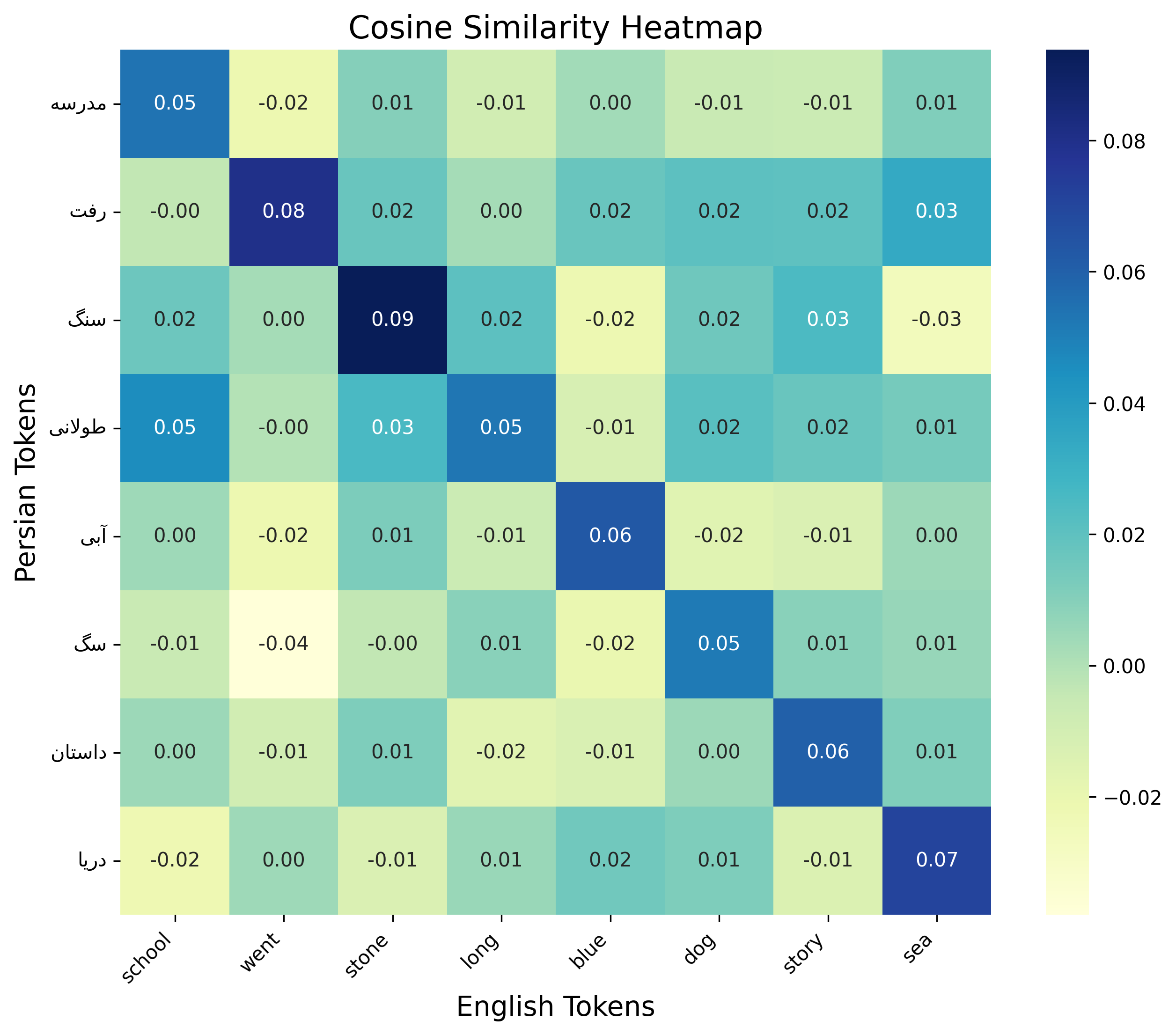}
\caption{Heatmap of Cosine similarity for English and Persian Equivalent Tokens. This heatmap highlights that tokens with similar meanings have higher similarity scores, emphasizing the effectiveness of the language warm-up process on Persian.}
\label{fig:cosine_similarity_heatmap}
\end{figure}

Furthermore, the alignment of embedding spaces for semantically equivalent English and Persian tokens was observed (see Figure \ref{fig:cosine_similarity_heatmap}). This result underscores the model’s capacity to harmonize semantic spaces across languages effectively.

These observations highlighted the success of the warm-up methodology in preparing the model for Persian-related tasks. The LoRA-adapted weights were eventually merged with the original model, producing a unified and improved version.

\subsection{Pre-training}

Pre-training with large and diverse datasets is essential for enabling effective language understanding and generation in LLMs \cite{Radford2018ImprovingLU, brown2020languagemodelsfewshotlearners}. This process allows models to learn complex statistical patterns \cite{inproceedings}, linguistic relationships \cite{Radford2018ImprovingLU, devlin2019bertpretrainingdeepbidirectional}, and, by extension, implicitly capture knowledge from the training corpus \cite{wei2022emergentabilitieslargelanguage}. This section outlines our pre-training methodology, beginning with the details of our dataset construction, followed by a comprehensive description of our data quality filtering methods, and concluding with an explanation of our training strategy for efficient model adaptation.

\subsubsection{Pre-training Dataset}

For pre-training, we used the Targoman Large Persian Corpus (TLPC) \cite{tlpc2024}, a large-scale Persian dataset designed for training large language models. The July 2024 release of TLPC included over 75 million documents and 41 billion tokens sourced from more than 800 diverse Persian websites. It covered a wide range of content, including news, blogs, forums, literature, Q\&A, legal, religious, medical, educational, and social media. This mix spans both formal and informal registers, offering a broad linguistic spectrum.

TLPC's scale and diversity exposed the model to the nuances and complexities of the Persian language, enhancing its robustness across various writing styles, topics, and domains, and enabling a more comprehensive language understanding. However, since TLPC was a raw dataset, we applied a rigorous filtering process (described later) to ensure its quality and suitability for model training.

In addition to TLPC, we augmented our pre-training corpus with the Persian Wikipedia dump (April 2024). As a source of high-quality, structured, and collaboratively curated content, Wikipedia helped the model learn general knowledge and improved its understanding of factual information in Persian.

\subsubsection{Dataset Pipeline}
Effective language model training requires high-quality data \cite{soldaini2024dolmaopencorpustrillion}, and raw text often contains noise, inconsistencies, and irrelevant content. This section details our data pipeline, designed to transform raw Persian text into a clean and consistent dataset. Our pipeline included content filtering, normalization, quality filtering, and deduplication—crucial steps for removing noise, standardizing the text, eliminating low-quality content, and ensuring data diversity. The following subsections explain each stage in the data transformation process.

\paragraph{Content Filtering}

Although the TLPC dataset encompasses various Persian news sources, many documents within these sources exhibit high similarity, thereby reducing their training value. To mitigate this issue, we selectively retain data from the YJC news agency, chosen for its broader coverage and greater diversity in reporting compared to other sources in TLPC. This approach reduces duplication while preserving a sufficient amount of news content for effective training.

To maintain a clear focus on the Persian language, we used FastText \cite{joulin2016bagtricksefficienttext} to filter out non-Persian documents, applying a language identification confidence threshold of 0.8.

Web-scraped data frequently contains harmful or toxic content \cite{luccioni-viviano-2021-whats}. Filtering such content is a common practice to reduce the risk of training models in toxic language \cite{Longpre2023APG, rae2022scalinglanguagemodelsmethods}. In our pipeline, we implemented a profanity filter that removes any document containing even a single instance of a term from a predefined list of swear words and offensive expressions.

After applying these filtering steps, we retain 12.4 million documents from the original TLPC for further processing.
 
\paragraph{Normalization}

Inconsistent writing styles in Persian and multiple character representations for the same concept pose significant challenges in processing the Persian language~\cite{shamsfard-etal-2010-step, oji2021parsinormpersiantoolkitspeech}. For instance, variations in the words ``\FR{کتابها}'' and ``\FR{کتاب‌ها}'' (both meaning ``books'') unnecessarily increase the tokenizer's vocabulary size and negatively impact model efficiency. These inconsistencies can also prevent the tokenizer from correctly identifying meaningful units and generating appropriate feature vectors for each word.

To address these challenges, we applied a \textbf{normalization process} that utilizes text processing algorithms primarily drawn from the \texttt{Hazm}~\cite{hazm} and \texttt{DadmaTool}~\cite{dadmatools} libraries, along with several custom implementations. This process improves data quality and model performance through the following key operations:

\begin{itemize}
    \item \textbf{Removing optional Arabic diacritics} to reduce noise;
    \item \textbf{Unifying character representations}, such as different forms of \emph{hamza}, or the letters ``\FR{ک}'' and ``\FR{ی}'', to reduce vocabulary size and ambiguity;
    \item \textbf{Applying Unicode replacements} for consistent phrase representation;
    \item \textbf{Eliminating irrelevant special characters}, such as control characters like \verb|\u200e| (left-to-right mark), to further clean the text.
\end{itemize}

Additionally, when processing Persian Wikipedia data, we removed sections such as \emph{Gallery}, \emph{References}, and \emph{External links}. While these sections are useful for human readers, they may distract the model from learning the core linguistic structure of the Persian language.

\paragraph{Quality Filtering}

Sources such as social media often contain impure or irrelevant information, including repetitive or incoherent user comments. Additionally, auto-generated content—such as excessive advertisements and SEO keyword stuffing—is common across Persian web data. These types of content can introduce misleading patterns, degrade model performance, and compromise the accuracy and reliability of downstream tasks. As a result, data refinement to eliminate low-quality information is a critical step in preparing datasets for LLM training~\cite{soldaini2024dolmaopencorpustrillion, rae2022scalinglanguagemodelsmethods}.

In this work, we explored heuristic techniques based on measurable text features to improve data quality for Persian LLMs. We adapted heuristics introduced in Gopher~\cite{rae2022scalinglanguagemodelsmethods} to accommodate the lexical characteristics of Persian and introduce additional criteria. For example, considering that approximately 65.6\% of Persian words contain six or fewer characters~\cite{persianlexicon}, we filterd out texts with atypical average word lengths. To further ensure textual coherence, we required the presence of common conjunctions and additive expressions such as ``\FR{و}'' (and), ``\FR{سپس}'' (then), and ``\FR{اینکه}'' (that). For a complete list of the filtering heuristics employed, please refer to Appendix~\ref{appendix:filtering_criteria}.

\paragraph{Repetitive Removal}
Repetitive text within a document is a strong indicator of low-quality data, often associated with uninformative or artificially generated content. Such repetition in training corpora has also been shown to exacerbate the well-documented tendency of language models to generate repetitive outputs~\cite{holtzman2020curiouscaseneuraltext}, ultimately degrading overall performance.To mitigate this, we employed techniques similar to those used in the curation of the Dolma corpus~\cite{soldaini2024dolmaopencorpustrillion}, detecting and eliminating redundancy at the line, phrase, and sentence levels. 

Following the application of quality filtering and repetition removal, we discarded approximately 3.5 million documents from the TLPC dataset and 700{,}000 documents from the Persian Wikipedia corpus.

\paragraph{Deduplication}

Deduplication is a critical preprocessing step for training large language models (LLMs), significantly enhancing training efficiency and reducing overfitting by removing redundant data~\cite{lee-etal-2022-deduplicating}. Prior research demonstrates that deduplication improves training dynamics by preventing models from memorizing repetitive content, thereby enabling a more focused and effective learning process~\cite{lee-etal-2022-deduplicating, hernandez2022scalinglawsinterpretabilitylearning}.

We performed deduplication using the MinHash algorithm~\cite{Minhash}, implemented via the \texttt{datatrove} library~\cite{datatrove}, which supports multi-core processing and is optimized for large-scale deduplication. MinHash constructs compact signatures for documents and compares them to efficiently detect near-duplicate instances while balancing computational cost and accuracy.

Our MinHash configuration consists of 10 buckets and 6 hash functions per bucket. To reduce computational overhead, we used 64-bit hash functions. Additionally, we employed bi-grams for signature generation, as they capture linguistic nuances and structural variations in Persian text more effectively than unigrams.

Using this setup, we removed approximately 2.8 million duplicate documents from the TLPC dataset, resulting in 6.1 million documents retained for training. To preserve the high-quality content of Wikipedia and minimize the risk of discarding valuable data, we do not apply deduplication to the Persian Wikipedia corpus.

\subsubsection{Training Explanation}

The following details the entire continual pre-training process, from the preparation of the final training dataset to the specific configurations and methodologies employed for fine-tuning the model.

\paragraph{Dataset Pre-processing and Tokenization}

We initially utilized an enhanced tokenizer that incorporates 5,000 additional Persian-specific tokens to efficiently process both English and Persian text. After tokenization, documents were segmented into chunks of 2,048 tokens each.

To ensure comprehensive data representation, we augmented the dataset through several strategies. First, we doubled the number of chunks from the Persian Wikipedia dataset to leverage its structured and knowledge-rich content. Additionally, we included 32,000 chunks from the translated Tiny-Stories dataset to preserve the model’s translation capabilities. These chunks—each containing 2,048 tokens—were then mixed to construct the final training dataset, promoting diverse exposure and reducing the risk of overfitting.

The final training dataset comprises approximately 2,100,000 chunks from the TLPC dataset, 182,000 chunks from Persian Wikipedia , and 32,000 chunks from the translated Tiny-Stories dataset. As each chunk contains 2,048 tokens, the total dataset size amounts to 4,739,072,000 tokens.

\paragraph{Training Configurations}\label{sec:training_configurations}

For efficient fine-tuning, we utilized LoRA (rank 64, $\alpha = $32) on attention and feed-forward layers, which were critical for language understanding and generation. To integrate Persian vocabulary and ensure coherent output, we additionally applied full fine-tuning to the embedding and output head layers. This results in approximately 200 million LoRA parameters and 120 million embedding and head parameters being updated during training, representing 8\% of the total model parameters.

Training efficiency was improved using Flash-Attention~\cite{dao2022flashattentionfastmemoryefficientexact} and a memory-optimized 8-bit AdamW optimizer (learning rate: $1\text{e}{-4}$, weight decay: $1\text{e}{-4}$, betas: 0.9 and 0.95). A cosine learning rate scheduler with a 250-step warm-up was used to ensure stable convergence.

To parallelize training across multiple GPUs, we employed Distributed Data Parallel (DDP), which enabled efficient scaling and synchronized updates across devices. We used a per-GPU batch size of 1, leveraging gradient accumulation with a step size of 64 to achieve an effective batch size of 64. Mixed-precision training with \texttt{bfloat16} balances speed and accuracy, while \texttt{TF32} is used for high-precision operations. Training was performed on two NVIDIA RTX 3090 GPUs, achieving a throughput of 5{,}000 tokens per second and lasting approximately 12 days.

\section{Supervised Fine-Tuning}

Supervised fine-tuning plays a vital role in adapting pre-trained language models for enhancing conversational abilities. This phase involved training the model on instruction-response pairs, enabling better adherence to user instructions and specialized task performance. For the Phi-3-mini-instruct model, which underwent initial alignment in its pre-training, reinforcement was necessary to restore its instruction-following capabilities after pre-training on Persian corpora, since continuous pre-training may have diminished these capabilities.

The dataset for supervised fine-tuning was constructed using a multi-faceted approach to ensure the model’s effectiveness in both Persian and English. Three key datasets were utilized:

\paragraph{Bactrian-X Dataset:}
The \textit{Bactrian-X} dataset \cite{li2023bactrianx} provided approximately 63,000 Persian instruction-response pairs, originally translated from English instructions with GPT3.5-generated responses.

\paragraph{Aya Dataset:}
To recover English capabilities, 50,000  samples were selected from the English portion of the \textit{Aya Dataset} \cite{singh2024aya}.

\paragraph{TED2020 Corpus:}
The \textit{TED2020 corpus} \cite{reimers-gurevych-2020-making} contributed 30,000 bilingual sentence pairs, formatted into instruction-response pairs. This dataset was intended to enhance the model’s bidirectional translation capabilities, fostering cross-lingual performance.




Supervised fine-tuning enhanced the model’s ability to follow instructions in Persian while helping the model to retain its English proficiency through careful training strategies. The loss function was tailored to focus solely on \textbf{Assistant} tokens. This approach contrasts with traditional language modeling, where the loss is computed across all tokens to build a generalized understanding of language. Optimization utilized LoRA with a rank of 32, modifying attention, feed-forward layers, and embedding weights for new tokens. Training employed a maximum context length of 512 tokens, a batch size of 4, and 64 gradient accumulation steps. The remaining hyperparameters were kept consistent with those used during the supervised tuning and pre-training step. 
At the conclusion of this phase, LoRA weights were merged with the model's base weights to produce the final model.

\section{Evaluation}
\subsection{Evaluation Datasets}

The evaluation dataset used in the Open Persian LLM Leaderboard\cite{openpersianllm2024} is designed to assess Persian language models across a wide range of linguistic and reasoning tasks. According to the publicly available information on the leaderboard’s official Hugging Face page, the dataset comprises five task categories: Part Multiple Choice, ARC Easy, ARC Challenge, MMLU Pro, and AUT Multiple Choice Persian. These tasks cover various domains such as general knowledge, logical reasoning, mathematics, and academic examinations, and collectively include over 40,000 Persian-language samples. The dataset features diverse input types—including plain text, mathematical expressions, and structured multiple-choice formats—and is evaluated using a standardized pipeline built on the LM Evaluation Harness. To prevent overfitting and preserve fairness, only a portion of the dataset is publicly accessible, and the internal details of the full evaluation set are not disclosed.

\subsection{Quantitative Evaluation}
This section evaluates the performance of our Persian LLM, Persian-Phi (based on Phi-3-mini), across several benchmarks and compares it to other state-of-the-art open-source Persian models based on the Llama family.

Our model is unique in that Phi-3-mini is a purely monolingual English model, requiring us to teach it Persian from scratch. Other leading models, such as the Dorna family (based on Llama 3), PersianMind, and MaralGPT (based on Llama 2) \cite{maralgpt2023maral7b}, leverage multilingual foundation models with a pre-existing, though varying, degree of knowledge across numerous languages, including Persian. This difference in starting point presents a unique challenge for our model, as we cannot pivot on existing cross-lingual knowledge within the model's architecture.

Quantitative evaluations, summarized in Table \ref{tab:model_comparison} ), were conducted by PartAI as an independent third party to ensure unbiased and accurate results. Our model demonstrates solid and competitive performance across all evaluated benchmarks (Part Multiple Choice, ARC, MMLU Pro\footnote{The model's performance on MMLU Pro was constrained by the limited context window (2,048 tokens CPT and 512 tokens SFT) used during adaptation, as this benchmark requires extended context understanding for multi-hop reasoning.}, and AUT Multiple Choice Persian). This result confirms the success of our curriculum-based approach in efficiently teaching Persian to an exclusively English-trained model.

\begin{table}[h]
\centering
\begin{tabular}{|l|c|c|c|c|c|c|}
\hline
\textbf{Model} & \textbf{\#Params (B)} & \textbf{Part MC
} & \textbf{ARC Easy} & \textbf{ARC Challenge} & \textbf{MMLU Pro} & \textbf{AUT MC} \\ \hline
PartAI Dorna2-8B & 8.03 & 35.52 & 75.28 & 53.52 & 24.1 & 53.45 \\ \hline
Meta-LLaMA3.1-8B & 8.03 & 36.68 & 78.4 & 60.4 & 21 & 54.24 \\ \hline
Gemma-2-2b-it & 2.61 & 31.12 & 71.26 & 57.72 & 16.23 & 49.9 \\ \hline
\textbf{Ours} & 3.85 & 30.56 & 64.65 & 51.00 & 17.18 & 43.98 \\ \hline
PersianMind-v1.0 & 6.82 & 29.27 & 58.91 & 48.32 & 15.51 & 45.36 \\ \hline
Maral-7B-alpha-1 & 7.24 & 26.67 & 44.54 & 32.88 & 15.99 & 36.09 \\ \hline
Phi-3-mini-4k-instruct & 3.82 & 27.37 & 36.78 & 36.78 & 17.89 & 35.1 \\ \hline
\end{tabular}
\caption{Comparison of different open-source models on the Open Persian LLM Leaderboard}
\label{tab:model_comparison}
\end{table}

We achieved strong results, outperforming several fine-tuned Persian models based on the Llama 2 architecture. Our model was outperformed only by Dorna-2, the current state-of-the-art model on this leaderboard. It is critical to note the difference in scale: Dorna-2 is built upon an $8 \text{B}$-parameter base model, utilizing twice the capacity of our $3.85 \text{B}$-parameter Persian-Phi. Despite this significant architectural difference, Persian-Phi achieves approximately $80\%$ of Dorna-2's aggregate performance. This result underscores the high efficiency of our cross-lingual adaptation strategy.

\section{Limitations}

While our approach demonstrates the efficacy of adapting lightweight monolingual models to low-resource languages, we acknowledge several limitations in our study.

First, due to computational constraints and our limited access to only two NVIDIA RTX 3090 GPUs for a short period of time, we were unable to conduct a full ablation study to empirically quantify the isolated impact of the warm-up stage. Ideally, a baseline model would be trained with identical hyperparameters but without the translation-based warm-up to rigorously benchmark the benefits of our curriculum learning approach. However, preliminary experiments with direct initialization indicated significant instability and slower convergence, suggesting that the warm-up phase is critical for aligning the new embedding space with the pre-trained English representations.

Finally, as with any cross-lingual transfer approach derived from an English-centric base model, there is a potential for "alignment bias," where the model’s reasoning patterns may reflect Western cultural norms rather than native Persian cultural nuances.

\section{Conclusion}

In this work, we presented \textbf{Persian-Phi}, a 3.8B parameter language model adapted from Microsoft’s Phi-3 Mini to support the Persian language. By implementing a novel curriculum learning pipeline—comprising tokenizer extension, embedding warm-up via bilingual narratives, and continual pre-training on filtered corpora—we successfully transferred the reasoning capabilities of a high-quality English monolingual model to a low-resource language.

Our findings challenge the prevailing assumption that robust multilingual support requires massive parameters or training from scratch. We demonstrated that with a strategic warm-up phase and parameter-efficient fine-tuning (PEFT), a compact model can achieve competitive performance on Persian benchmarks. This approach not only democratizes access to LLM technology for the Persian-speaking community but also provides a scalable, resource-efficient blueprint for adapting monolingual models to other underrepresented languages. Future work will focus on scaling this methodology to larger architectures and further refining the context window limitations to enhance long-text reasoning.

\section*{Acknowledgments}
We extend our sincere gratitude to the \textbf{Part AI Research Center} for their generous provision of the GPU resources that made this research possible.

\bibliographystyle{unsrt}  
\bibliography{references}  

\appendix

\section{Detailed Quality Filtering Criteria} \label{appendix:filtering_criteria}

This appendix details the heuristic criteria used to filter the dataset for training the Persian LLM.  These heuristics aim to remove low-quality content and ensure the dataset contains relevant and well-formed text.

The following criteria were applied:

\begin{enumerate}
    \item Documents must contain between 50 and 20,000 words.  This range balances sufficient content with avoiding excessively short or long, potentially unstructured, texts.
    \item Based on the observation that approximately 65.6\% of Persian words have six or fewer characters\cite{persianlexicon}, we filter out documents with an average word length outside the 3-to-7-character range.  This criterion targets texts with unusual character combinations or non-standard writing.
    \item Documents where the ratio of special symbols (e.g., \#, \ldots) to words exceeds 0.1 are discarded. This filter identifies texts that prioritize formatting over meaningful content.
    \item At least 80\% of the words in a document must contain at least one Persian letter.  This criterion excludes content primarily in other languages or composed mainly of symbols.
    
    \item Documents with more than 90\% of lines starting with a bullet point are removed, as they likely represent lists rather than cohesive narratives.
    
    \item Documents with more than 30\% of lines ending with an ellipsis are removed, suggesting incomplete thoughts or a lack of clarity.

    \item Each document must contain at least two "necessary words," such as conjunctions and additive words like \FR{" و "} (and), \FR{" سپس "} (then), \FR{" اینکه "} (That). This criterion aims to filter out poorly constructed sentences, low-quality machine-translated text, or bot-generated content.
    \item Documents with a line-to-word ratio greater than 0.1 are discarded.  This filter targets texts with very short lines and excessive spacing often found in poetry, simple lists, or SEO-optimized content.
\end{enumerate}

By systematically applying these criteria, we aim to create a refined dataset suitable for training effective Persian language models.

\section{A Sample of Bilingual Story Generated by the Warmed-up Model}

\begin{table}[h!]
\centering
\begin{tabular}{|p{15cm}|} 
\hline
\textbf{Prompt:} Once upon a time, there were two friends who have similar interests.\\
\hline
\textbf{Response:} \\ \hline

Once upon a time, there were two friends who have similar interests. They both loved to play in the park. One day, they decided to go to the park together.\\
\FR{یک بار دو دوست بودند که جاه طلبی مشابهی داشتند. هر دو دوست داشتند در پارک بازی کنند. یک روز تصمیم گرفتند با هم به پارک بروند.} \\ \hline

When they arrived, they saw a big tree. They both wanted to climb it. But one of them was scared and said, "I don't want to climb the tree."\\
\FR{وقتی رسیدند، درخت بزرگی را دیدند. هر دو می خواستند از آن بالا بروند. اما یکی از آنها ترسید و گفت: '' من نمی خواهم از درخت بالا بروم.''} \\ \hline

The other friend said, "Come on, it's fun! I'll help you."\\
\FR{دوست دیگر گفت: '' بیا، جالب است! من به تو کمک خواهم کرد.''} \\ \hline

The scared friend was still scared, but the other friend said, "Don't worry, I'll be with you."\\
\FR{دوست ترسیده هنوز می ترسید، اما دوست دیگر گفت:'' نگران نباش، من با تو خواهم بود.''} \\ \hline

So the two friends climbed the tree together. They had a lot of fun and laughed a lot.\\
\FR{بنابراین دو دوست با هم از درخت بالا رفتند. خیلی خوش گذشت و خندیدند.} \\ \hline

When they were done, they both said, "That was fun! Let's do it again!"\\
\FR{وقتی کارشان تمام شد، هر دو گفتند: ''این جالب بود! بیا دوباره این کار را انجام دهیم!''} \\ \hline

The two friends were happy and they hugged each other. They were glad they had the same interests and were able to do something together.\\
\FR{دو دوست خوشحال شدند و همدیگر را در آغوش گرفتند. آنها خوشحال بودند که جاه طلبی مشابهی داشتند و می توانستند با هم کاری انجام دهند.} \\
\hline
\end{tabular}
\label{tab:story_generation_example}

\caption{An example of bilingual story generation by the model. This example showcases the model's ability to correctly use Persian grammar and vocabulary, which results from the warm-up process. The coherent and contextually appropriate Persian text indicates that the model has effectively learned to align and utilize Persian linguistic structures by performing simple translation tasks.}
\end{table}

\end{document}